\documentclass[a4paper]{article}

\usepackage{INTERSPEECH2021}

\usepackage{graphicx}
\usepackage{algorithm}
\usepackage{algorithmic}
\usepackage{subcaption}
\usepackage{booktabs}
\usepackage[inline]{enumitem}
\setlist[enumerate,1]{label=\textbf{(\alph*)}}

\title{Zero-Shot Federated Learning with New Classes for Audio Classification}
\name{Gautham Krishna Gudur$^1$, Satheesh Kumar Perepu$^2$}
\address{
  $^1$Global AI Accelerator, Ericsson\\
  $^2$Ericsson Research}
\email{gautham.krishna.gudur@ericsson.com, perepu.satheesh.kumar@ericsson.com}

\begin{document}

\maketitle

\sloppy

\begin{abstract}

Federated learning is an effective way of extracting insights from different user devices while preserving the privacy of users. However, new classes with completely unseen data distributions can stream across any device in a federated learning setting, whose data cannot be accessed by the global server or other users. To this end, we propose a unified zero-shot framework to handle these aforementioned challenges during federated learning. We simulate two scenarios here -- 1) when the new class labels are not reported by the user, the traditional FL setting is used; 2) when new class labels are reported by the user, we synthesize \textit{Anonymized Data Impressions} by calculating class similarity matrices corresponding to each device's new classes followed by unsupervised clustering to distinguish between new classes across different users. Moreover, our proposed framework can also handle statistical heterogeneities in both labels and models across the participating users. We empirically evaluate our framework on-device across different communication rounds (FL iterations) with new classes in both local and global updates, along with heterogeneous labels and models, on two widely used audio classification applications -- keyword spotting and urban sound classification, and observe an average deterministic accuracy increase of $\sim$4.041\% and $\sim$4.258\% respectively.

\end{abstract}

\section{Introduction}

\label{label:introduction}

Deep learning for audio classification is a broad research area with applications like Keyword Spotting (KWS), urban sound identification, etc. KWS is an important application for detecting keywords of importance to specific users, which could be used as voice commands to on-device personal assistants such as Amazon's Alexa, Apple's Siri, etc. \cite{cite:keywordspotting}. Urban environment sound classification is another interesting application particularly in context-aware computing, urban informatics \cite{cite:urbansounds}. The emergence of deep neural networks have conveniently alleviated problems of creating shallow (hand-picked) features to achieve state-of-the-art performance in such acoustic classification tasks \cite{cite:dnn_speech,cite:large_audio_cnn}. With the recent compute capabilities vested in resource-constrained devices, there is a huge research focus on audio classification using on-device deep learning \cite{cite:kws_smallfootprint,cite:kwscnn_smallfootprint}.

\begin{figure}[ht]
    \centering
    \includegraphics[width=180pt, height=160pt]{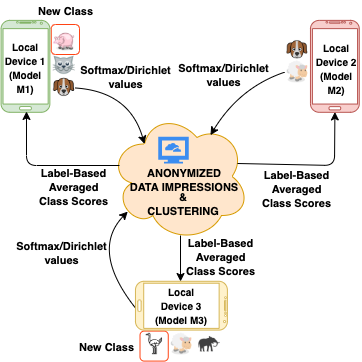}
    \caption{Architecture of proposed Federated Learning framework with new classes streaming in across different users.}
    \label{fig:newclass_block_diagram}
    %\vskip -0.1in
\end{figure}

Such applications require characterization of insights across numerous user devices for personalization, and collaborative on-device deep learning becomes necessary. Federated Learning (FL) is a decentralized method of training neural networks by securely sharing model updates with a server without the need to transfer sensitive local user data \cite{cite:federated_learning,cite:fedavg}. On-device federated learning has been an active area of research addressing challenges on secure communication protocols, optimization, privacy-preserving networks, etc. \cite{cite:challenges_methods,cite:fed_framework}. However, handling new/unseen classes in local devices and training them in an FL setting for the global model to possess characteristics of the new classes is a challenging task, since data transfer from local device to server and vice versa is not feasible. Moreover, the new class information of one user is not known among the other users as well, hence the new classes could be similar or different between the users. In addition, there are multiple statistical heterogeneities like model heterogeneities (ability of end-users to architect their own local models), label heterogeneities and non-IIDness across various communication rounds/FL iterations (disparate data and label distributions across devices).

One way of handling model heterogeneities and independence in a federated learning setting is by using knowledge distillation \cite{cite:knowledge_distillation} with a common student model architecture on each local device \cite{cite:FedMD}. Label and model heterogeneities are handled in an inertial Human Activity Recognition scenario in \cite{cite:dlhar_ijcai}. Federated learning for keyword spotting \cite{cite:fl_kws}, and new class learning and identification in various speech recognition settings are addressed in \cite{cite:new_class1, cite:new_class2}. \cite{cite:new_class3} proposes a new augmentation technique to reduce false reject rates and addresses algorithmic constraints in FL-KWS training to label examples with no visibility. However, the scope of our proposed work is different in the nature that it primarily addresses identification and similarity detection of new labels in a zero-shot manner when heterogeneous label and model distributions exist across various FL iterations and users. To the best of our knowledge, none of the papers discuss new label identification in FL settings with statistical heterogeneities for audio classification.

Our scientific contributions are:
\begin{enumerate*}[label=\textbf{(\arabic*)}]
    \item A framework with zero-shot learning mechanism by synthesizing \textit{Anonymized Data Impressions} from class similarity matrices to identify new classes for keyword spotting and urban sound detection in on-device FL settings.
    \item Provide two scenarios for label acquisition -- when class label is reported by user, and when class label is not, and propose unsupervised clustering to identify and differentiate between newly reported classes.
    \item Handling statistical heterogeneities such as heterogeneous distributions in labels, data and models across devices and FL iterations.
\end{enumerate*}

\section{Our Approach}
\label{section:our_approach}

In this section, we discuss the problem formulation of new classes in FL, and our proposed framework (Algorithm \ref{alg:proposed_method}) to handle the same. The overall architecture is given in Figure \ref{fig:newclass_block_diagram}.

\subsection{Problem Formulation}
We assume the following scenario in federated learning. Suppose there are $M$ nodes (devices) in the FL network, holding private local data $\mathcal{D}_i = \{x_{i,j},y_{i,j}\}$ where $i$ is the FL iteration and $j$ is the user index. Each node consists of public data $\mathcal{D}_0=\{x_0,y_0\}$. The public data is assumed to be present across the global and all local users as discussed in \cite{cite:FedMD} to handle the various statistical (model) heterogeneities which is a common phenomena in FL. The overall label-set of public dataset is $Y = \{y_0\}$, which represent its unique labels. We re-purpose this public dataset as test set and do not expose it to local models during FL training iterations, but expose only during testing for consistency. Our work's main contribution is to propose a framework to identify new labels across different users without transferring private data in FL setting. We also assume each user can stream data with new labels at any iteration which does not belong to public label-set $Y$, i.e. $\mathbf{y}_{i,j} \not\in Y$. In other words, the global user has no idea of these new labels.

\begin{algorithm}[ht]
   \caption{Our Proposed Framework}
   \label{alg:proposed_method}
   \centering
   \begin{algorithmic}
   \STATE \textbf{Input:} Public Dataset $\mathcal{D}_0\{x_0,y_0\}$, Private Datasets $\mathcal{D}_m^i$, Total users $M$, Total iterations $I$, LabelSet $l_m$ for each user, Overall Public LabelSet $Y$,\\
   \STATE \textbf{Output:} Trained Model scores $f_G^I$
   %\REPEAT
   \vspace{0.1cm}
   \STATE Initialize $f_G^0 = \mathbf{0}$ (Global Model Scores)
   \vspace{0.1cm}
   \FOR{$i=1$ \textbf{to} $I$}
   \FOR{$m=1$ \textbf{to} $M$}
   \vspace{0.1cm}
   \STATE \textbf{Build:} Model $\mathcal{D}_m^i$ and predict $f_{\mathcal{D}_m^i}(x_0)$
   \vspace{0.1cm}
   \STATE \textbf{Local Update:}
   \STATE \textbf{Choice 1: New classes are not reported}\\
   $f_{\mathcal{D}_m^i}(x_0) = f_G^I(x_0^{l_m})+\alpha f_{\mathcal{D}_m^i}(x_0)$, where $f_G^I(x_0^{l_m})$ are global scores of $l_m$ with $m^{th}$ user, $\alpha = \frac{len(\mathcal{D}_m^i)}{len(\mathcal{D}_0)}$
   \vspace{0.1cm}
   \STATE \textbf{Choice 2: New classes are reported}\\
   Train a new model with $\mathcal{D}_0$ and $\mathcal{D}_m^i$ (new data) together, and send weights of the last layer ($\mathbf{W}_m^i$) to global user.
   %\vspace{0.1cm}
   \ENDFOR
   \vspace{0.1cm}
   \STATE \textbf{Global Update:}
   \STATE \textbf{Choice 1: No user reports new classes}\\
   Update label wise \\
   $f_G^{i+1} = \displaystyle \sum_{m=1}^{M}\beta_m f_{\mathcal{D}_m^i}(x_0)$, where \\
   $\beta = 
   \begin{cases}
   1 & \text{If labels are unique} \\
   \text{acc}(f_{\mathcal{D}_m^{i+1}}(x_0)) & \text{if labels are not unique}
   \end{cases}$\\
   \vspace{0.1cm}
   where $\text{acc}(f_{\mathcal{D}_m^{i+1}}(x_0))$ is the accuracy metric, defined by the ratio of correctly classified samples to total samples for a given local model.\\
   \vspace{0.1cm}
   \STATE \textbf{Choice 2: Any user reports new classes}\\
   Create \textit{Data Impressions (DI)} for each user $m$ with weights $\mathbf{W}_m^i$ (Section \ref{section:anonymized_data_impressions}). Average \textit{DI} of all users with new classes, $\mathbf{X}^i=\sum_{m \in M_{S_k}}\mathbf{X}_m^i$, where $M_{S_k}$ is set of users with new label $k$.\\
   Perform \textit{k-medoids clustering} on $\mathbf{X}^i$ across $M_{S_k}$. Number of clusters = Number of new labels ($l_{new}$).\\
   \vspace{0.1cm}
   Update public dataset with new DI ($\mathbf{X}^i$), $\mathcal{D}_{new} = \mathcal{D}_0 \bigcup \mathbf{X}^i$, add $l_{new}$ to $l_{m}$ and $Y$.
   \ENDFOR
\end{algorithmic}
\end{algorithm}

\subsection{Anonymized Data Impressions}
\label{section:anonymized_data_impressions}

The main challenge/objective is to detect similar labels across different users in FL heterogeneous settings without the knowledge of local user data. This necessitates us to construct anonymized data without transferring raw sensitive data, and identify new class similarities on the anonymized data. We motivate our framework from the creation of Data Impressions (DI) using zero-shot learning as proposed in \cite{cite:zeroshot} to compute \textit{Anonymized Data Impressions}. Let us assume a model $\mathcal{M}$ with input $\mathbf{X}$ and output $\mathbf{y}$, where $\mathbf{X} \in \mathcal{R}^{M\times N}$ is the set of features and $\mathbf{y}\in \mathcal{R}^M$. Now, the anonymized feature set $\bar{\mathbf{X}}$, which has same properties of $\mathbf{X}$, can be synthesized in two steps:

\begin{table*}[ht]
\vskip -0.05in
\caption{Model Architectures (filters in each layer), Labels and Audio frames per FL iteration across user devices for both datasets. Note the disparate model architectures and labels across users.}
\label{table:models_labels_iterations}
\centering
\resizebox{\textwidth}{!}{
\begin{tabular}{|c|c|c|c|c|}
\toprule
\textbf{}           & \textbf{User\_1}      & \textbf{User\_2}      & \textbf{User\_3}      & \textbf{Global\_User}\\ 
\midrule
\textbf{Architecture}                                                                       & \begin{tabular}[c]{@{}c@{}}2-Layer CNN\\ (16, 32)\\ Softmax Activation\end{tabular}           & \begin{tabular}[c]{@{}c@{}}3-Layer CNN\\ (16, 16, 32)\\ ReLU Activation\end{tabular}          & \begin{tabular}[c]{@{}c@{}}3-Layer Depth-Separable CNN\\ (16, 16, 32)\\ ReLU Activation\end{tabular}          &  –\\ 
\hline
\textbf{Keywords} & \{Yes, No, Up, Down\}         & \{Up, Down, Left, Right\}       & \{Left, Right, On, Off\}   & 
\begin{tabular}[c]{@{}c@{}}\{Yes, No, Up, Down, Left, Right,\\ Left, Right, On, Off\} \end{tabular}                       \\ 
\hline
\textbf{\begin{tabular}[c]{@{}c@{}}Keyword Frames\\ per iteration \end{tabular}}           & \begin{tabular}[c]{@{}c@{}}\{200-300, 200-300,\\ 200-300, 200-300\}\\ \end{tabular}                             & \begin{tabular}[c]{@{}c@{}}\{200-300, 200-300,\\ 200-300, 200-300\}\\ \end{tabular}                             & \begin{tabular}[c]{@{}c@{}}\{200-300, 200-300,\\ 200-300, 200-300\}\\ \end{tabular}                             & \begin{tabular}[c]{@{}c@{}}\{300*8\} = 2400\end{tabular}\\
\hline
\textbf{Sounds} & \begin{tabular}[c]{@{}c@{}}\{air conditioner, car horn,\\ children playing\}\end{tabular}         & \begin{tabular}[c]{@{}c@{}}\{children playing, dog bark, \\drilling \}\end{tabular}       & \begin{tabular}[c]{@{}c@{}}\{drilling, engine idling, \\gun shot, jackhammer\}\end{tabular}   & 
\begin{tabular}[c]{@{}c@{}}\{air conditioner, car horn, children playing,\\ dog bark, drilling, engine idling, gun shot, jackhammer\} \end{tabular}                       \\ 
\hline
\textbf{\begin{tabular}[c]{@{}c@{}}Sound Frames\\ per iteration \end{tabular}}           & \begin{tabular}[c]{@{}c@{}}\{40-50, 40-50, 40-50\}\\ \end{tabular}                             & \begin{tabular}[c]{@{}c@{}}\{40-50, 40-50, 40-50\}\\ \end{tabular}                             & \begin{tabular}[c]{@{}c@{}}\{40-50, 40-50,\\ 40-50, 40-50\}\\ \end{tabular}                             & \begin{tabular}[c]{@{}c@{}}\{50*8\} = 400\end{tabular}\\
\bottomrule
\end{tabular}
}
\vskip -0.1in
\end{table*}

\textbf{(a) Sample Softmax Values:}
The first step is to sample the softmax values from the Dirichlet distribution  \cite{cite:Dirchlet}. The Class Similarity Matrix (\textit{CSM}) is created which contains important information on how similar the classes are to each other. If the classes are similar, we expect the softmax values are concentrated over these labels. \textit{CSM} is obtained by considering the weights of the model's last layer. Typically, any classification model has the final layer as fully-connected layer with a softmax non-linearity. If the classes are similar, we find similar weights between connections of the penultimate layer to the nodes of the classes \cite{cite:zeroshot}. The Class Similarity Matrix is constructed as, 
\begin{align}
    C(i,j) = \frac{\mathbf{w}_i^T\mathbf{w}_j}{||\mathbf{w}_i||||\mathbf{w}_j||}
\end{align}
where $\mathbf{w}_i$ is the vector of weights connecting the previous layer nodes to the class node $i$. $\mathbf{C} \in \mathcal{R^{K\times K}}$ is the Class Similarity Matrix for $K$ classes. We then sample the softmax values as,
\begin{align}
    \text{Softmax} = Dir(K,C)
\end{align}
where $C$ is concentration parameter which controls the spread of softmax values over class labels.

\textbf{(b) Creating Anonymized Data Impressions:}
Let $\mathbf{Y}^k = [\mathbf{y}_1^k,\mathbf{y}_2^k,\cdots,\mathbf{y}_N^k] \in \mathcal{R}^{K\times N}$ be the N softmax vectors corresponding to class $k$, sampled from Dirichlet distribution from previous step. Once we obtain the softmax values, we compute the synthesized data features (Data Impressions) by solving the following optimization problem using model $\mathcal{M}$ and sampled softmax values $\mathbf{Y}^k$
\begin{align}
    \bar{\mathbf{x}} = \arg \underset{\mathbf{x}}{\min} L_{CE} (\mathbf{y}_i^k,\mathcal{M}(\mathbf{x}))
\end{align}
To solve this optimization problem, we initialize the input $\mathbf{x}$ to be random input and iterate until cross-entropy loss ($L_{CE}$) minimization. This process is repeated for all $K$ categories. In this way, anonymized data impressions are created for each class without the visibility of original input data. We use the TensorFlow framework \cite{cite:tensorflow} for all our experiments.

\begin{figure*}[ht]
%\vskip -0.05in
\centering
    \begin{subfigure}[b]{0.249\textwidth}
        \includegraphics[width=\linewidth]{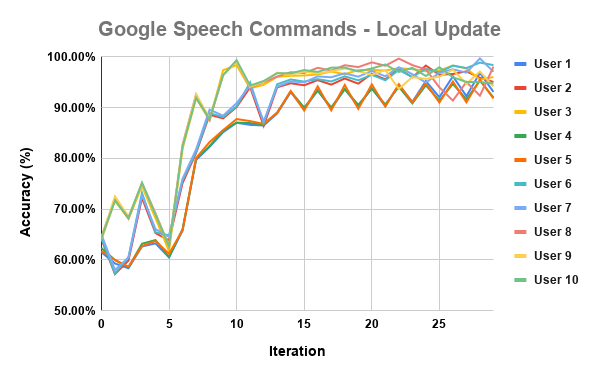}
        \caption{GKWS - Local Update}
        \label{fig:google_local}
    \end{subfigure}%
    \begin{subfigure}[b]{0.249\textwidth}
        \includegraphics[width=\linewidth]{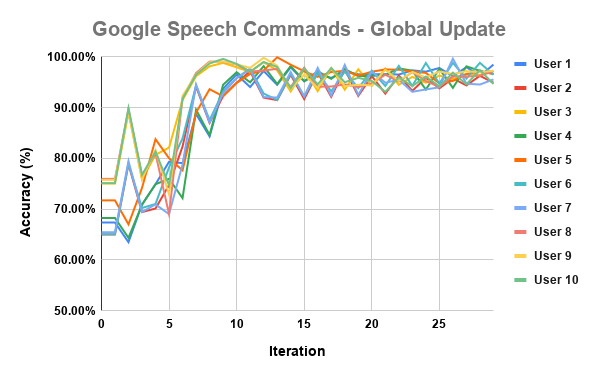}
        \caption{GKWS - Global Update}
        \label{fig:google_global}
    \end{subfigure}%
    \begin{subfigure}[b]{0.249\textwidth}
        \includegraphics[width=\linewidth]{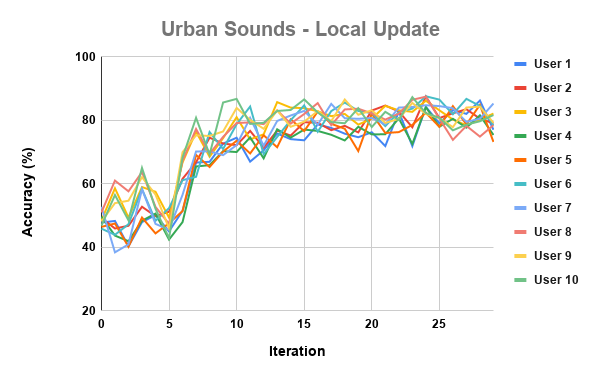}
        \caption{US8K - Local Update}
        \label{fig:urban_local}
    \end{subfigure}%
    \begin{subfigure}[b]{0.249\textwidth}
        \includegraphics[width=\linewidth]{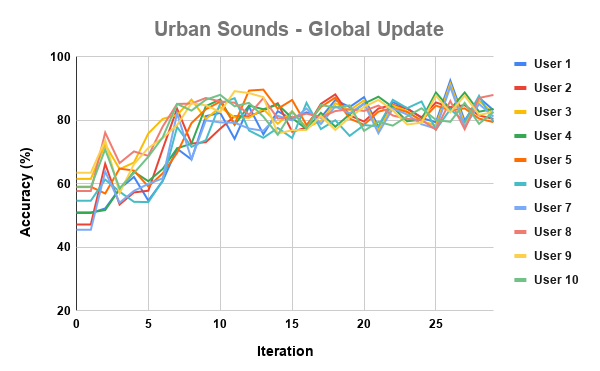}
        \caption{US8K - Global Update}
        \label{fig:urban_global}
    \end{subfigure}%
\caption{Local-Global update accuracies across 10 users and 30 FL iterations for both datasets with new classes and heterogeneities.}
\label{fig:hetero_figs}
%\vskip -0.1in
\end{figure*}

%\vskip -0.5in
\subsection{Proposed Framework}
There are three steps in our proposed framework (Algorithm \ref{alg:proposed_method}).

    \textbf{(a) Build}: Each local user creates their own model with their local private data for a specific iteration.
    
    \textbf{(b) Local Update}: In this step, if new classes are not reported, we perform simple weighted $\alpha$-update \cite{cite:aiot20}, where $\alpha$ governs the contributions of new and old models across FL iterations shown in Algorithm \ref{alg:proposed_method} Choice 1. If new classes are reported, we train the new class data along with public dataset, and send the new model weights to global user (Choice 2).
    
    \textbf{(c) Global update}: In this step, if no user reports new classes, we perform label-based averaging using the parameter $\beta$, which governs contributions of overlapping labels using corresponding test accuracies (Choice 1). If user reports new classes, we create \textit{Anonymized Data Impressions (DI)} for new classes followed by unsupervised clustering using k-medoids with motivations from \cite{cite:clustering} (Choice 2).
    
Typically, statistical heterogeneities are widely observed in practical FL settings, hence Choice 1 handles heterogeneities in local and global update steps \cite{cite:dlhar_ijcai}, while Choice 2 handles new classes in our proposed framework.

\section{Experiments and Results}
\label{section:exp_results}

We simulate our experiments using \textit{Raspberry Pi 2} as our user device with \textbf{Google Speech Commands (GKWS)} \cite{cite:google_keywords} and \textbf{UrbanSound8K (US8K)} \cite{cite:urbansounds} datasets across 10 FL iterations/communication rounds using our proposed framework. In GKWS, we choose the keywords: Yes, No, Up, Down, Left, Right, On, Off, Stop and Go, and perform regular Mel-frequency Cepstral Coefficients (MFCC) extraction as performed in \cite{cite:keywordspotting}, with sampling frequency of 14400 HZ. The MFCC data is divided into 20 windows and each window is of size 50 ms. US8K, an environmental sound dataset, consists of 10 classes of sound events: air conditioner, car horn, children playing, dog bark, drilling, engine idling, gun shot, jackhammer, siren and street music. We perform similar preprocessing as performed in GKWS for US8K as well.

\textbf{Public Dataset:}
We create a Public Dataset ($D_0$) with 2400 audio frames for GKWS (8 keywords with 300 frames each), and 400 audio frames for US8K (8 sounds with 50 frames each) as shown in Table \ref{table:models_labels_iterations}. $D_0$ is visible to both global and local users in each FL iteration, and is updated with data synthesized for unseen/new classes only -- Anonymized Data Impressions.

We initially consider eight labels with an initial Public Dataset in both datasets before streaming new classes (Table \ref{table:models_labels_iterations}). We simulate two scenarios for testing just our zero-shot framework – 1) new classes only (homogeneous) with limited users and FL iterations (3 users and 10 FL iterations) for effective analysis of results, 2) new classes with statistical heterogeneities in both labels and models as performed in \cite{cite:dlhar_ijcai}, (10 users and 30 FL iterations). This exhibits near-real-time model heterogeneities as shown in Table \ref{table:changing_models}, and effective convergence.

\textbf{New Classes:} We introduce two new/unseen labels \{Stop, Go\} for GKWS and \{Siren, Street music\} for US8K across four FL iterations and two users. In the homogeneous case, for GKWS, we induce 400 samples each with Stop class in iteration 4 for both User 1 and User 2, and 500 samples each with Stop in User 1 iteration 8 and Go in User 2 iteration 8. Similarly, we induce 50 samples each with Siren class in iteration 4 for both User 1 and User 2, and 50 samples each with Siren in User 1 iteration 8 and Street music in User 2 iteration 8. This is the FL scenario with new classes without any heterogeneities. We also discuss similar FL scenarios with statistical heterogeneities.

\begin{table}[ht]
\vskip -0.05in
\caption{Details of heterogeneities - model architectures (filters) and new classes changing across FL iterations and users for both datasets.}
\label{table:changing_models}
\centering
\resizebox{7.4cm}{!}{%
\begin{tabular}{ccc}
\toprule
\textbf{Iteration}      & \textbf{New Model}    & \textbf{New Class}    \\
\midrule
User 1 Iteration 6      & \begin{tabular}[c]{@{}c@{}}3-Layer ANN (16, 16, 32)\\ ReLU Activation\end{tabular} &  - \\
\hline
User 1 Iteration 8      & \begin{tabular}[c]{@{}c@{}}1-Layer CNN (16)\\ Softmax Activation\end{tabular} &  - \\
\hline
User 2 Iteration 4, 6   & \begin{tabular}[c]{@{}c@{}}3-Layer CNN (16, 16, 32)\\ Softmax activation\end{tabular} & Stop/Siren  \\
\hline
User 3 Iteration 5      & \begin{tabular}[c]{@{}c@{}}4-Layer CNN (8, 16, 16, 32)\\ Softmax activation\end{tabular} &  -      \\
\hline
User 4 Iteration 3, 7   &  -  & Go/Street Music    \\
\hline
User 6 Iteration 3, 5   &  -  & Stop/Siren   \\
\hline
User 9 Iteration 4      &  -  & Stop/Siren   \\
\bottomrule
\end{tabular}}
\vskip -0.05in
\end{table}

\textbf{(a) Label Heterogeneities:} 
In every FL iteration, we consider a random number of audio frames generated between 200-300 samples per label for GKWS, while 40-50 samples per label for US8K. We split these labels across three users such that labels can either be unique or overlapping across users. We also simulate non-IIDness across FL iterations with disparities in both labels and distributions in data (\textit{statistical heterogeneities}).

\textbf{(b) Model Heterogeneities:}
We consider the three model architectures as shown in Table \ref{table:models_labels_iterations} motivated from \cite{cite:keywordspotting, cite:depthwise}, and also change model architectures, filters and activation functions across FL iterations in addition to label heterogeneities with new classes (Table \ref{table:changing_models}). The user iterations are chosen at random.

\begin{figure*}[ht]
\centering
    \begin{subfigure}[b]{0.249\textwidth}
        \includegraphics[width=\linewidth]{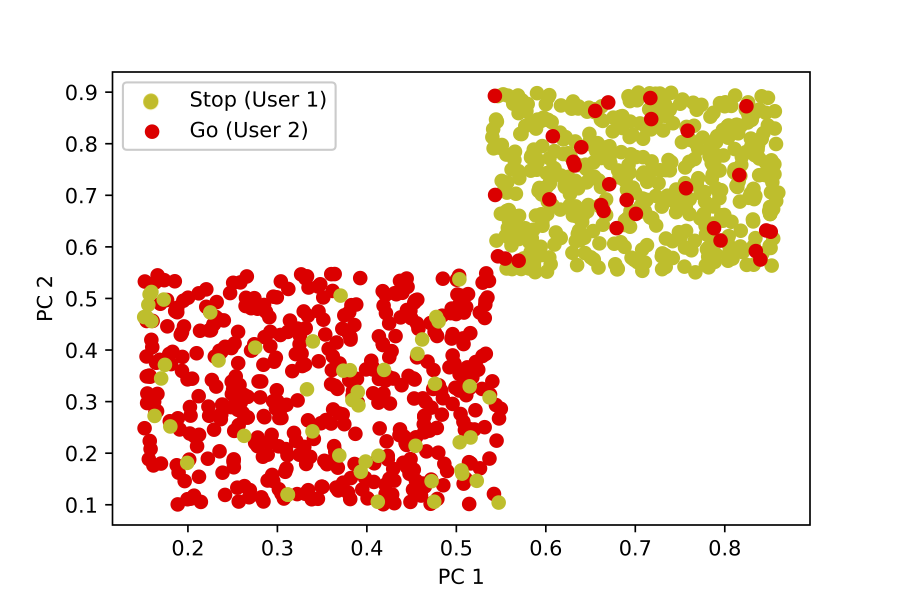}
        \caption{GKWS - PCA - Different Class}
        \label{fig:differentclass_kws}
    \end{subfigure}%
    \begin{subfigure}[b]{0.249\textwidth}
        \includegraphics[width=\linewidth]{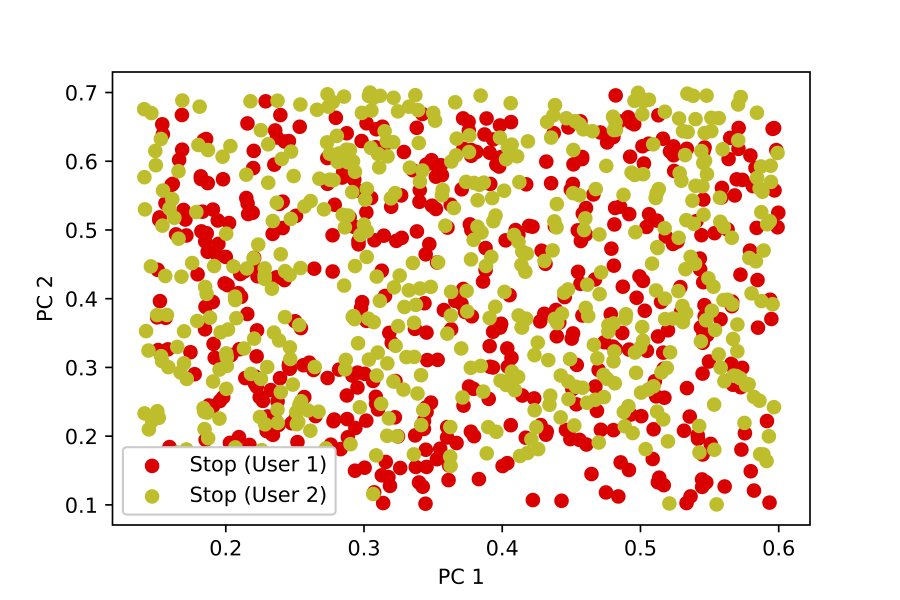}
        \caption{GKWS - PCA - Same Class}
        \label{fig:sameclass_kws}
    \end{subfigure}%
    \begin{subfigure}[b]{0.249\textwidth}
        \includegraphics[width=\linewidth]{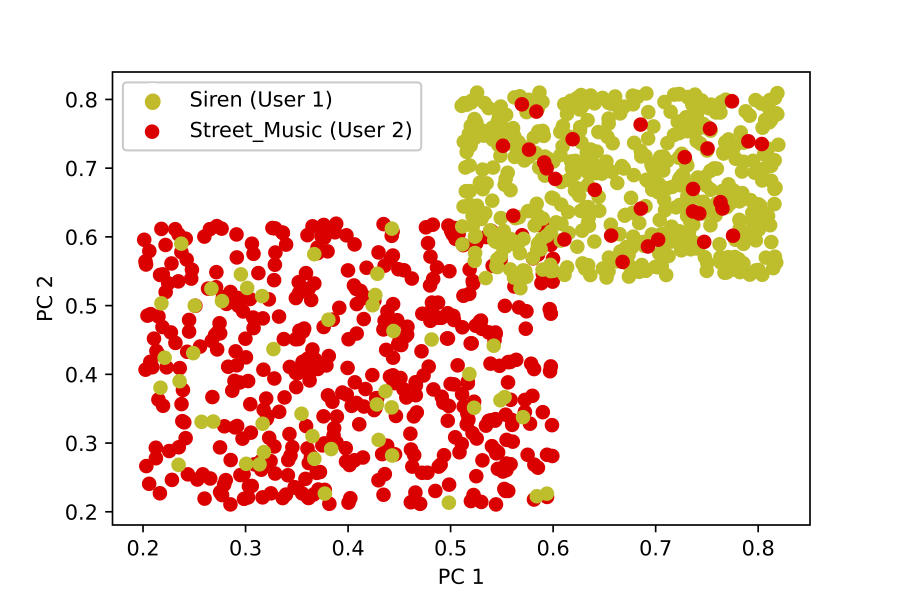}
        \caption{US8K - PCA - Different Class}
        \label{fig:differentclass_urban}
    \end{subfigure}%
    \begin{subfigure}[b]{0.249\textwidth}
        \includegraphics[width=\linewidth]{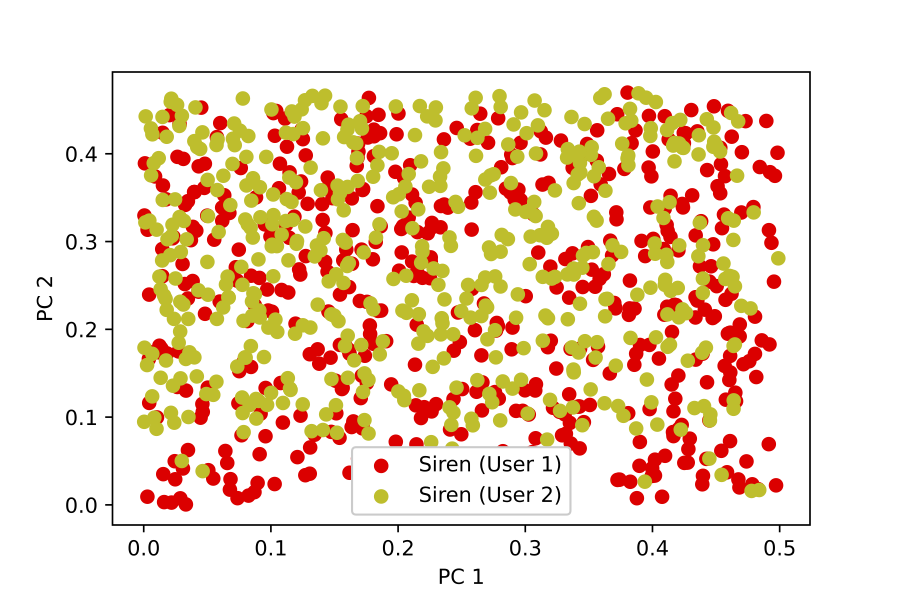}
        \caption{US8K - PCA - Same Class}
        \label{fig:sameclass_urban}
    \end{subfigure}%
\caption{PCA (with 2 dimensions) of k-medoids unsupervised clustering with new classes, with same and different classes for both datasets.}
\label{fig:clustering_results}
\end{figure*}

\subsection{Discussion on Results}
\label{section:discussion_results}

\begin{table}[ht]
\vskip -0.05in
\caption{Local-global update accuracies (\%) for both datasets across 3 users with just new classes, without heterogeneities.}
\label{table:homo_results}
\centering
\resizebox{\linewidth}{!}{
\begin{tabular}{ccccccc}
\toprule
\textbf{}        & \multicolumn{3}{c}{\textbf{GKWS}}                    & \multicolumn{3}{c}{\textbf{US8K}}                    \\
\midrule
\textbf{User}             & \textbf{Local} & \textbf{Global} & \textbf{Increase} & \textbf{Local} & \textbf{Global} & \textbf{Increase} \\ \hline
User 1          & 89.684         & 93.166          & 3.482             & 76.526          & 80.214           & 3.688             \\
User 2          & 91.888         & 95.28           & 3.391             & 75.272          & 77.944            & 2.672             \\
User 3          & 91.517         & 94.727          & 3.211             & 77.61          & 81.838           & 4.228             \\
\textbf{Average} & \textbf{91.03} & \textbf{94.391} & \textbf{3.361}    & \textbf{76.469} & \textbf{80}  & \textbf{3.529}    \\
\bottomrule
\end{tabular}
}
\vskip -0.1in
\end{table}

\begin{figure}[ht]
\vskip -0.05in
%\centering
%\resizebox{8cm}{!}{
%\begin{table}
    \begin{minipage}[b]{0.45\linewidth}
        \centering
        %\resizebox{4cm}{!}{%
        \includegraphics[width=\textwidth]{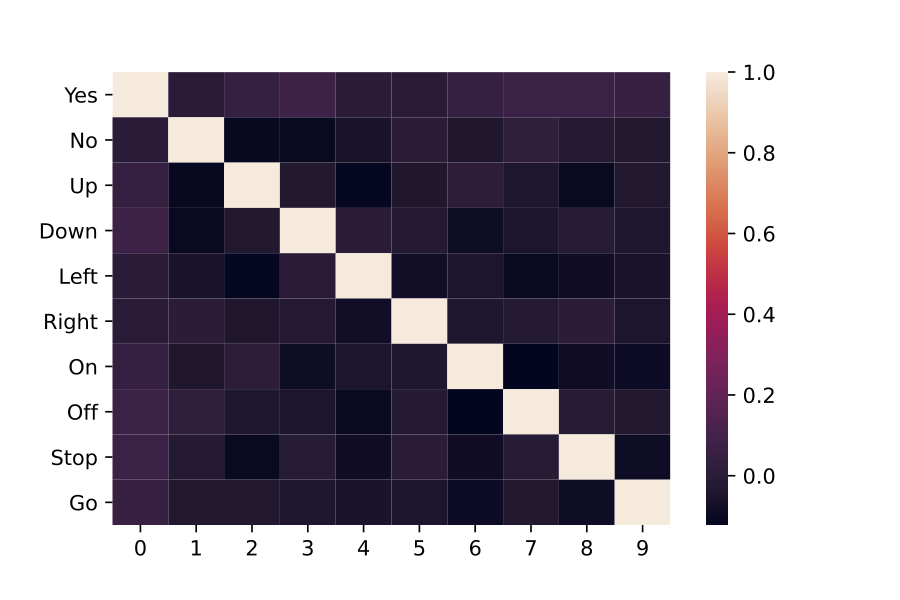}%}
        \caption{Class Similarity Matrix for GKWS.}
        \label{fig:classsimilarity_google}
    \end{minipage}%
    \hspace{0.5cm}
    \begin{minipage}[b]{0.45\linewidth}
        \caption{Final accuracies (\%) for 10 users, 30 FL iters.}
        \label{table:hetero_results}
        \centering
        \resizebox{\textwidth}{!}{%
        \begin{tabular}{lll}
        \toprule
        \textbf{Update}         &   \textbf{GKWS}       &   \textbf{US8K}   \\
        \midrule
        \textbf{Local}          &   92.5                &   78.24           \\
        \textbf{Global}         &   96.541              &   82.498          \\
        \textbf{Increase}       & \textbf{4.041}        & \textbf{4.258}    \\
        \bottomrule
        \end{tabular}}
    \end{minipage}
\vskip -0.1in
\end{figure}

From Table \ref{table:homo_results}, we can observe that there is an accuracy increase in the FL scenario with just new classes (without heterogeneities) in corresponding global updates for all three users than their respective local update accuracies for both datasets in spite of new classes streaming in. The average local-global accuracy increase across all 10 FL iterations and 3 users is $\sim$3.361\% and $\sim$3.529\% respectively for GKWS and US8K. Similarly, we can also observe that with our proposed framework, the final global accuracies (with convergence after all FL iterations) even with new classes and heterogeneities are 96.541\% and 82.498\% (Figure \ref{table:hetero_results}) which are much higher than their respective local update accuracies. The corresponding local-global update accuracies across 30 FL iterations and 10 users are shown in Figure \ref{fig:hetero_figs}. The class similarity matrix of different classes for GKWS is showcased in Figure \ref{fig:classsimilarity_google}, which elucidates the misclassifications. We can also infer that the clusters effectively formed with k-medoids are equal to the number of new classes, which are visualized using Principal Component Analysis (PCA) in two-dimensions. The new classes can either be different or same across user devices (Figure \ref{fig:clustering_results}), and these classes are correctly mapped back to the respective end-user devices. The new labels are then finally added to the overall label set while the corresponding averaged data impressions are added to the public dataset.

\subsection{On-Device Performance}
\label{section:ondevice}

Raspberry Pi 2 (900MHz quad-core ARM Cortex-A7 CPU with 1GB RAM) is used for evaluating our proposed FL framework as it has similar hardware and software (HW/SW) specifications to predominant contemporary IoT/mobile devices. The computation times are identical for both datasets due to similar preprocessing. The size of the models used are also 520 kB, 350 kB, 270 kB respectively for user architectures mentioned in Table \ref{table:models_labels_iterations}.

\begin{table}[ht]
\vskip -0.05in
\caption{Computation Times with Raspberry Pi 2}
\label{table:timeOnPi}
\centering
\begin{tabular}{lll}
\toprule
\textbf{Process}    & \textbf{Time}   \\
\midrule
\begin{tabular}[c]{@{}c@{}}Training time per epoch\\ in an FL iteration ($i$)\end{tabular}  & $\sim$1.2 sec     \\
\hline
Inference time                                                                              & $\sim$11 ms       \\
\bottomrule
\end{tabular}
\vskip -0.1in
\end{table}

\section{Conclusions}
\label{section:conclusion}

This paper presents a novel framework for handling new labels in a federated learning setting. We propose a zero-shot learning framework by synthesizing Anonymized Data Impressions from Class Similarity matrices to learn new classes across different user devices. We also account for heterogeneities in labels and models across different FL communication rounds, and systematically analyze the results for two widely used audio classification applications -- keyword spotting and urban sound classification. We further demonstrate the effectiveness and scalability of our proposed FL framework by simulating our experiments on-device using a Raspberry Pi 2.

\bibliographystyle{IEEEtran}

\bibliography{References}

\end{document}